\documentstyle[latex-acl]{article}

\title{GRAMMAR SPECIALIZATION THROUGH ENTROPY THRESHOLDS}
\author{Christer Samuelsson\\
Swedish Institute of Computer Science\\
Box 1263 S-164 28 Kista, Sweden\\
Internet: {\tt christer@sics.se}}

\def\myln{\mbox{ln }}
\def\E{\mbox{ E}}

\def\twosum#1#2{\mathrel{\sum\limits_{#1}\limits^{#2}}}
\def\limtox#1#2{\mathrel{\lim\limits_{#1\to#2}}}
\def\limtoinf#1{\limtox{#1}{\infty}}

\begin{document}
\maketitle

\section{Abstract}

Explanation-based generalization is used to extract a specialized grammar
from the original one using a training corpus of parse trees. This allows
very much faster parsing and gives a lower error rate, at the price of a
small loss in coverage. Previously, it has been necessary to specify the
tree-cutting criteria (or operationality criteria) manually; here they are
derived automatically from the training set and the desired coverage of the
specialized grammar.
This is done by assigning an entropy value to each node in the parse trees
and cutting in the nodes with sufficiently high entropy values.

\section{BACKGROUND}

Previous work by Manny Rayner and the author, see \cite{Samuelsson+Rayner:91}
attempts to tailor an existing natural-language
system to a specific application domain by extracting a specialized grammar
from the original one using a large set of training examples. The training
set is a treebank consisting of implicit parse trees that each specify a
verified analysis of an input sentence.
The parse trees are implicit in the sense that each node in the tree is the
(mnemonic) name of the grammar rule resolved on at that point, rather than
the syntactic category of the LHS of the grammar rule as is the case in an
ordinary parse tree. Figure~\ref{FigTrees} shows five examples of implicit
parse trees.  The analyses are verified in the sense that each analysis
has been judged to be the preferred one for that input sentence by a human
evaluator using a semi-automatic evaluation method.

A new grammar is created by cutting up each implicit parse tree in the
treebank at appropriate points, creating a set of new rules that consist
of chunks of original grammar rules.
The LHS of each new rule will be the LHS phrase of the original grammar
rule at the root of the tree chunk and the RHS will be the RHS phrases
of the rules in the leaves of the tree chunk. For example, cutting up the
first parse tree of Figure~\ref{FigTrees} at the {\it NP\/} of the rule
\verb!vp_v_np! yields rules 2 and 3 of Figure~\ref{FigSpecRules}.

The idea behind this is to create a specialized grammar that retains a
high coverage but allows very much faster parsing. This has turned out to
be possible --- speedups compared to using the original grammar of in
median 60 times were achieved at a cost in coverage of
about ten percent, see \cite{Samuelsson:94a}.\footnote{Other more easily
obtainable publications about this are in preparation.}
Another benefit from the method is a decreased error rate when the system
is required to select a preferred analysis.
In these experiments the scheme was applied to the grammar of a version of
the SRI Core Language Engine \cite{AlshawiEd:92} adapted to the Atis domain
for a speech-translation task \cite{RaynerEA:93} and large corpora of real
user data collected using Wizard-of-Oz simulation. The resulting specialized
grammar was compiled into LR parsing tables, and a special LR parser exploited
their special properties, see \cite{Samuelsson:94b}.

The technical vehicle previously used to extract the specialized grammar
is explanation-based generalization (EBG), see e.g.~\cite{MitchellEA:86}.
Very briefly, this consists of redoing the derivation of each training
example top-down by letting the implicit parse tree drive a rule expansion
process, and aborting the expansion of the specialized rule currently being
extracted if the current node of the implicit parse tree meets a set of
tree-cutting criteria\footnote{These are usually referred to as
``operationality criteria'' in the EBG literature.}. In this case the
extraction process is invoked recursively to extract subrules rooted in the
current node. The tree-cutting criteria can be local (``The LHS of the
original grammar rule is an {\it NP\/},'') or dependent on the rest of the
parse tree (``that doesn't dominate the empty string only,'') and previous
choices of nodes to cut at (``and there is no cut above the current node
that is also labelled {\it NP\/}.'').

A problem not fully explored yet is how to arrive at an optimal choice of
tree-cutting criteria. In the previous scheme, these must be specified
manually, and the choice is left to the designer's intuitions.
This article addresses the problem of automating this process and presents
a method where the nodes to cut at are selected automatically using the
information-theoretical concept of entropy.
Entropy is well-known from physics, but the concept of perplexity is perhaps
better known in the speech-recognition and natural-language communities.
For this reason, we will review the concept of entropy at this point, and
discuss its relation to perplexity.

\subsection{Entropy}

Entropy is a measure of disorder. Assume for example that a physical system
can be in any of $N$ states, and that it will be in state $s_i$ with
probability $p_i$. The entropy $S$ of that system is then
\begin{eqnarray*}
S = \sum_{i=1}^N - \: p_i \cdot \myln p_i
\end{eqnarray*}
If each state has equal probability, i.e.~if $p_i = \frac{1}{N}$ for
all $i$, then
\begin{eqnarray*}
S = \sum_{i=1}^N -\frac{1}{N} \cdot \myln \frac{1}{N} = \myln N
\end{eqnarray*}
In this case the entropy is simply the logarithm of the number of states
the system can be in.

To take a linguistic example, assume that we are trying to predict the next
word in a word string from the previous ones. Let the next word be $w_k$
and the previous word string $w_1,...,w_{k-1}$. Assume further that we have
a language model that estimates the probability of each possible next word
(conditional on the previous word string). Let these probabilities be $p_i$
for $i = 1,...,N$ for the $N$ possible next words $w_k^i$,
i.e.~$p_i = p(w_k^i \mid w_1,...,w_{k-1})$.
The entropy is then a measure of how hard this prediction problem is:
\begin{eqnarray*}
\lefteqn{S(w_1,...,w_{k-1}) = }\\
& &  \twosum{i=1}{N} - \:
p(w_k^i \mid w_1,...,w_{k-1}) \cdot \myln p(w_k^i \mid w_1,...,w_{k-1})
\end{eqnarray*}
If all words have equal probability, the entropy is the logarithm of the
branching factor at this point in the input string.

\subsection{Perplexity}

Perplexity is related to entropy as follows.
The observed perplexity $P_o$ of a language model with respect
to an (imaginary) infinite test sequence $w_1,w_2,...$ is defined through
the formula (see \cite{Jelinek:90})
\begin{eqnarray*}
\myln P_o = \limtoinf{n} - \frac{1}{n} \myln p(w_1,...,w_n)
\end{eqnarray*}
Here $p(w_1,...,w_n)$ denotes the probability of the word string $w_1,...,w_n$.

Since we cannot experimentally measure infinite limits, we terminate after
a finite test string $w_1,...,w_M$, arriving at the measured perplexity
$P_m$:
\begin{eqnarray*}
\myln P_m = - \frac{1}{M} \myln p(w_1,...,w_M)
\end{eqnarray*}

Rewriting $p(w_1,...,w_k)$ as
$p(w_k \mid w_1,...,w_{k-1}) \cdot p(w_1,...,w_{k-1})$ gives us
\begin{eqnarray*}
\myln P_m = \frac{1}{M} \twosum{k=1}{M} - \myln p(w_k \mid w_1,...,w_{k-1})
\end{eqnarray*}

Let us call the exponential of the expectation value of
$- \myln p(w \mid {\it String})$ the local perplexity $P_l({\it String})$,
which can be used as a measure of the
information content of the initial {\it String\/}.
\begin{eqnarray*}
\lefteqn{\myln P_l(w_1,...,w_{k-1}) =
\E (- \myln p(\xi_k \mid w_1,...,w_{k-1})) = } \\
 & & \twosum{i=1}{N} - \: p(w_k^i \mid w_1,...,w_{k-1}) \cdot
\myln p(w_k^i \mid w_1,...,w_{k-1})
\end{eqnarray*}

Here $\E(\eta)$ is the expectation value of $\eta$ and the summation is
carried out over all $N$ possible next  words $w_k^i$. Comparing this
with the last equation of the previous section, we see that this is
precisely the entropy $S$ at point $k$ in the input string.
Thus, the entropy is the logarithm of the local perplexity at a given
point in the word string. If all words are equally probable, then the
local perplexity is simply the branching factor at this point. If the
probabilities differ, the local perplexity can be viewed as a
generalized branching factor that takes this into account.

\subsection{Tree entropy}

We now turn to the task of calculating the entropy of a node in a parse
tree. This can be done in many different ways; we will only describe two
different ones here.

Consider the small test and training sets of Figure~\ref{FigTrees}.
\begin{figure}
{\small
\begin{verbatim}
Training examples:

                         s_np_vp
                           /\
                    np_pron  vp_v_np
                       |        /\
     s_np_vp          lex      /  \
       /\              |      /    \
np_pron  vp_v_np       I   lex      np_np_pp
   |        /\              |           /\
  lex    lex  np_det_n    need  np_det_n  pp_prep_np
   |      |      /\                /\          /\
   I    want  lex  lex          lex  lex    lex  lex
               |    |            |    |      |    |
               a  ticket         a  flight  to  Boston

                           s_np_vp
                              /\
                             /  \
     s_np_vp         np_det_n    vp_vp_pp
       /\               /\          /\
np_pron  vp_v_np     lex  lex   vp_v  pp_prep_np
   |        /\         |    |      |        /\
  lex      /  \       The flight  lex    lex  np_num
   |      /    \                   |      |     |
  We   lex      np_np_pp        departs  at    lex
        |          /\                           |
      have        /  \                         ten
          np_det_n    pp_prep_np
             /\            /\
          lex  lex      lex  np_det_n
           |    |        |      /\
           a  departure  |   lex  lex
                        in    |    |
                             the morning

Test example:

    s_np_vp
       /\
np_pron  vp_v_np
   |        /\
  lex      /  \
   |      /    \
  He   lex      np_np_pp
        |           /\
     booked        /  \
           np_det_n    pp_prep_np
               /\           /\
            lex  lex       /  \
             |    |       /    \
             a  ticket lex      np_np_pp
                        |          /\
                        |  np_det_n  pp_prep_np
                       for    /\         /\
                           lex  lex   lex  lex
                            |    |     |    |
                            a  flight to  Dallas
\end{verbatim}
} 
\caption{A tiny training set}
\label{FigTrees}
\end{figure}
Assume that we wish to calculate the entropy of the phrases of the
rule {\it PP $\rightarrow$ Prep NP\/}, which is named \verb!pp_prep_np!.
In the training set, the LHS {\it PP\/} is attached to the RHS {\it PP\/}
of the rule \verb!np_np_pp! in two cases and to the RHS {\it PP\/} of the
rule \verb!vp_vp_pp! in one case, giving it the entropy $ - \frac{2}{3}
\mbox{ln} \frac{2}{3} - \frac{1}{3} \mbox{ln} \frac{1}{3} \approx 0.64$.
The RHS preposition {\it Prep\/} is always a lexical lookup, and the entropy
is thus zero\footnote{Since there is only one alternative, namely a lexical
lookup. In fact, the scheme could easily be extended to encompass including
lexical lookups of particular words into the specialized rules by
distinguishing lexical lookups of different words; the entropy would then
determine whether or not to cut in a node corresponding to a lookup,
just as for any other node, as is described in the following.},
while the RHS {\it NP\/} in one case attaches to the LHS of rule
\verb!np_det_np!, in one case to the LHS of rule \verb!np_num!, and in one
case is a lexical lookup, and the resulting entropy is thus
$ - \mbox{ln} \frac{1}{3} \approx 1.10$.

The complete table is given here:
\begin{quote}
\begin{tabular}{l|rrrrr}
Rule			&LHS	& 1st RHS	&2nd RHS\\
\hline
\verb!s_np_vp!		&0.00	&0.56		&0.56\\
\verb!np_np_pp!		&0.00	&0.00 		&0.00\\
\verb!np_det_n!		&1.33	&0.00		&0.00\\
\verb!np_pron!		&0.00	&0.00		&---\\
\verb!np_num!		&0.00	&0.00		&---\\
\verb!vp_vp_pp!		&0.00	&0.00		&0.00\\
\verb!vp_v_np!		&0.00	&0.00		&0.64\\
\verb!vp_v!		&0.00	&0.00		&---\\
\verb!pp_prep_np!	&0.64	&0.00		&1.10\\
\end{tabular}
\end{quote}

If we want to calculate the entropy of a particular node in a parse
tree, we can either simply use the phrase entropy of the RHS node, or
take the sum of the entropies of the two phrases that are unified in
this node. For example, the entropy when the RHS {\it NP\/} of the rule
\verb!pp_prep_np! is unified with the LHS of the rule \verb!np_det_n!
will in the former case be $1.10$ and in the latter case be
$1.10 + 1.33 = 2.43$.

\section{SCHEME OVERVIEW}

In the following scheme, the desired coverage of the specialized grammar is
prescribed, and the parse trees are cut up at appropriate places without
having to specify the tree-cutting criteria manually:
\begin{enumerate}
\item
Index the treebank in an and-or tree where the or-nodes correspond to
alternative choices of grammar rules to expand with and the and-nodes
correspond to the RHS phrases of each grammar rule.
Cutting up the parse trees will involve selecting a set of or-nodes in the
and-or tree. Let us call these nodes ``cutnodes''.
\item
Calculate the entropy of each or-node. We will cut at each node whose
entropy exceeds a threshold value. The rationale for this is that we wish
to cut up the parse trees where we can expect a lot of variation i.e.~where
it is difficult to predict which rule will be resolved on next. This
corresponds exactly to the nodes in the and-or tree that exhibit high
entropy values.
\item
The nodes of the and-or tree must be partitioned into equivalence classes
dependent on the choice of cutnodes in order to avoid redundant derivations
at parse time.\footnote{
This can most easily be seen as follows: Imagine two identical, but
different portions of the and-or tree. If the roots and leaves of
these portions are all selected as cutnodes, but the distribution of
cutnodes within them differ, then we will introduce multiple ways of
deriving the portions of the parse trees that match any of these two
portions of the and-or tree.}
Thus, selecting some particular node as a cutnode may cause other nodes to
also become cutnodes, even though their entropies are not above the threshold.
\item
Determine a threshold entropy that yields the desired coverage.
This can be done using for example interval bisection.
\item
Cut up the training examples by matching them against the and-or
tree and cutting at the determined cutnodes.
\end{enumerate}

It is interesting to note that a textbook method for {\it constructing\/}
decision trees for classification from attribute-value pairs is to
minimize the (weighted average of the) remaining entropy\footnote{
Defined slightly differently, as described below.} over all
possible choices of root attribute, see \cite{Quinlan:86}.

\section{DETAILED SCHEME}

First, the treebank is partitioned into a training set and a test set. The
training set will be indexed in an and-or tree and used to extract the
specialized rules. The test set will be used to check the coverage of the
set of extracted rules.

\subsection{Indexing the treebank}

Then, the set of implicit parse trees is stored in an and-or tree. The parse
trees have the general form of a rule identifier \verb!Id! dominating a list
of subtrees or a word of the training sentence. From the current or-node of
the and-or tree there will be arcs labelled with rule identifiers
corresponding to previously stored parse trees.
{}From this or-node we follow an arc labelled \verb!Id!, or add a new one if
there is none. We then reach (or add) an and-node indicating the RHS phrases
of the grammar rule named \verb!Id!.
Here we follow each arc leading out from this and-node in turn to
accommodate all the subtrees in the list. Each such arc leads to an or-node.
We have now reached a point of recursion and can index the corresponding
subtree. The recursion terminates if \verb!Id! is the special rule identifier
\verb!lex! and thus dominates a word of the training sentence, rather than
a list of subtrees.

\begin{figure}
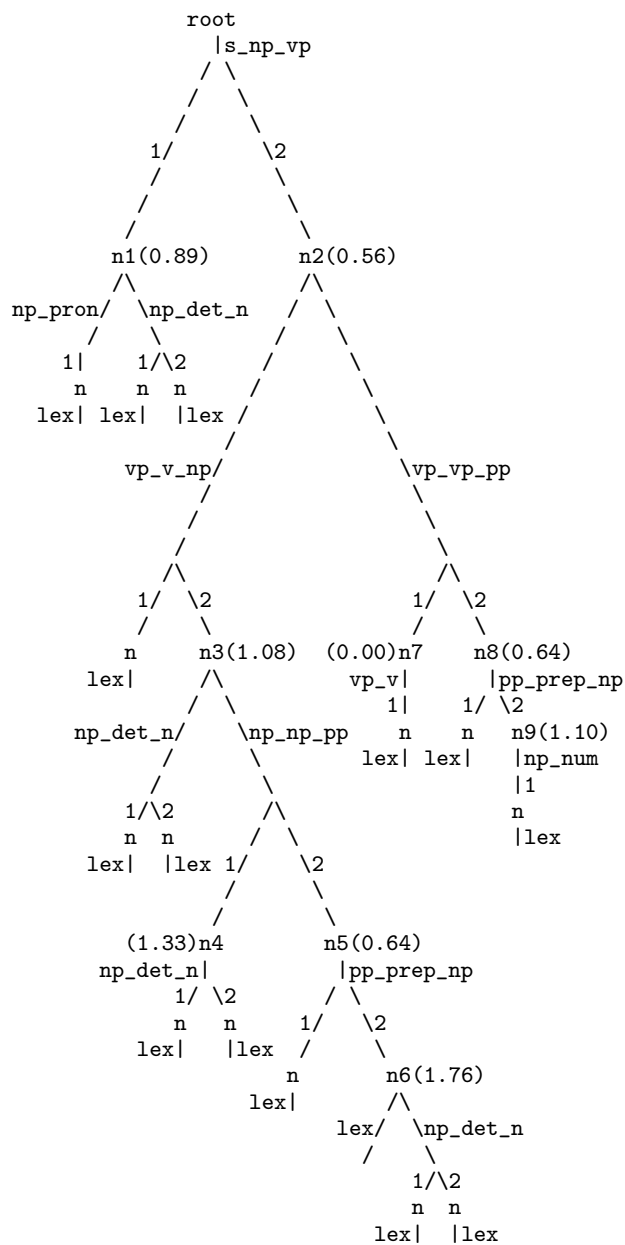

{\small
\begin{verbatim}
              root
                |s_np_vp
               / \
              /   \
             /     \
           1/       \2
           /         \
          /           \
         /             \
        n1(0.89)       n2(0.56)
        /\             /\
np_pron/  \np_det_n   /  \
      /    \         /    \
    1|    1/\2      /      \
     n    n  n     /        \
  lex| lex|  |lex /          \
                 /            \
         vp_v_np/              \vp_vp_pp
               /                \
              /                  \
             /                    \
            /\                    /\
          1/  \2                1/  \2
          /    \                /    \
         n     n3(1.08)  (0.00)n7    n8(0.64)
      lex|     /\          vp_v|      |pp_prep_np
              /  \            1|    1/ \2
     np_det_n/    \np_np_pp    n    n   n9(1.10)
            /      \        lex| lex|   |np_num
           /        \                   |1
         1/\2       /\                  n
         n  n      /  \                 |lex
      lex|  |lex 1/    \2
                 /      \
                /        \
         (1.33)n4        n5(0.64)
       np_det_n|          |pp_prep_np
             1/ \2       / \
             n   n     1/   \2
          lex|   |lex  /     \
                      n       n6(1.76)
                   lex|       /\
                          lex/  \np_det_n
                            /    \
                                1/\2
                                n  n
                             lex|  |lex
\end{verbatim}
} 
\caption{The resulting and-or tree}
\label{FigAndOr}
\end{figure}

Indexing the four training examples of Figure~\ref{FigTrees} will result in
the and-or tree of Figure~\ref{FigAndOr}.

\subsection{Finding the cutnodes}

Next, we find the set of nodes whose entropies exceed a threshold value.
First we need to calculate the entropy of each or-node. We will here
describe three different ways of doing this, but there are many others.
Before doing this, though, we will discuss the question of redundancy
in the resulting set of specialized rules.

We must equate the cutnodes that correspond to the same type of phrase.
This means that if we cut at a node corresponding to e.g.~an {\it NP\/},
i.e.~where the arcs incident from it are labelled with grammar rules whose
left-hand-sides are {\it NP\/}s, we must allow {\em all\/} specialized
{\it NP\/} rules to be potentially applicable at this point, not just the
ones that are rooted in this node. This requires that we by transitivity
equate the nodes that are dominated by a cutnode in a structurally equivalent
way; if there is a path from a cutnode $c_1$ to a node $n_1$ and a path from
a cutnode $c_2$ to a node $n_2$ with an identical sequence of labels, the
two nodes $n_1$ and $n_2$ must be equated. Now if $n_1$ is a cutnode, then
$n_2$ must also be a cutnode even if it has a low entropy value.
The following iterative scheme accomplishes this:
\begin{quote}
{\bf Function} $N^*(N^0)$
\begin{enumerate}
\item $i := 0$;
\item {\bf Repeat} $i := i+1$; $N^i := N(N^{i-1})$;
\item {\bf Until} $N^i = N^{i-1}$
\item {\bf Return} $N^i$;
\end{enumerate}
\end{quote}
Here $N(N^j)$ is the set of cutnodes $N^j$ augmented with those induced in
one step by selecting $N^j$ as the set of cutnodes. In practice this was
accomplished by compiling an and-or graph from the and-or tree and the set
of selected cutnodes, where each set of equated nodes constituted a vertex
of the graph, and traversing it.

In the simplest scheme for calculating the entropy of an or-node, only the RHS
phrase of the parent rule, i.e.~the dominating and-node, contributes to the
entropy, and there is in fact no need to employ an and-or tree at all, since
the tree-cutting criterion becomes local to the parse tree being cut up.

In a slightly more elaborate scheme, we sum over the entropies of the nodes
of the parse trees that match this node of the and-or tree. However, instead
of letting each daughter node contribute with the full entropy of the LHS
phrase of the corresponding grammar rule, these entropies are weighted with
the relative frequency of use of each alternative choice of grammar rule.

For example, the entropy of node \verb!n3! of the and-or tree of
Figure~\ref{FigAndOr} will be calculated as follows:
The mother rule \verb!vp_v_np! will contribute the entropy associated
with the RHS {\it NP\/}, which is, referring to the table above, 0.64.
There are 2 choices of rules to resolve on, namely \verb!np_det_n! and
\verb!np_np_pp! with relative frequencies $\frac{1}{3}$ and $\frac{2}{3}$
respectively. Again referring to the entropy table above, we find that
the LHS phrases of these rules have entropy 1.33 and 0.00 respectively.
This results in the following entropy for node \verb!n3!:
\begin{eqnarray*}
S(n_3) = 0.64 + \frac{1}{3} \cdot 1.33 + \frac{2}{3} \cdot 0.00 = 1.08
\end{eqnarray*}

The following function determines the set of cutnodes $N$ that either
exceed the entropy threshold, or are induced by structural equivalence:
\begin{quote}
{\bf Function} $N(S_{min})$
\begin{enumerate}
\item $N := \{n : S(n) > S_{min} \}$;
\item {\bf Return} $N^*(N)$;
\end{enumerate}
\end{quote}
Here $S(n)$ is the entropy of node $n$.

In a third version of the scheme, the relative frequencies of the daughters
of the or-nodes are used directly to calculate the node entropy:
\begin{eqnarray*}
S(n) = \sum_{n_i : \langle n,n_i \rangle \in A} - \:
p(n_i|n) \cdot \mbox{ ln }p(n_i|n)
\end{eqnarray*}
Here $A$ is the set of arcs, and $\langle n,n_i \rangle$ is an arc from $n$
to $n_i$. This is basically the entropy used in \cite{Quinlan:86}.
Unfortunately, this tends to promote daughters of cutnodes to in turn become
cutnodes, and also results in a problem with instability, especially in
conjunction with the additional constraints discussed in a later section,
since the entropy of each node is now dependent on the choice of cutnodes.
We must redefine the function $N(S)$ accordingly:
\newpage
\begin{quote}
{\bf Function} $N(S_{min})$
\begin{enumerate}
\item $N^0 := \emptyset$;
\item {\bf Repeat} $i := i+1$; \newline
      $N := \{n : S(n|N^{i-1}) > S_{min} \}$; $N^i := N^*(N)$;
\item {\bf Until} $N^i = N^{i-1}$
\item {\bf Return} $N^i$;
\end{enumerate}
\end{quote}
Here $S(n|N^j)$ is the entropy of node $n$ given that the set of cutnodes
is $N^j$. Convergence can be ensured\footnote{albeit in exponential time}
by modifying the termination criterion to be
\begin{quote}
3. {\bf Until} $\exists j \in [0,i-1] : \rho(N^i,N^j) < \delta(N^i,N^j)$
\end{quote}
for some appropriate set metric $\rho(N_1,N_2)$ (e.g.~the size of the
symmetric difference) and norm-like function $\delta(N_1,N_2)$ (e.g.~ten
percent of the sum of the sizes), but this is to little avail, since we
are not interested in solutions far away from the initial
assignment of cutnodes.

\subsection{Finding the threshold}

We will use a simple interval-bisection technique for finding the appropriate
threshold value. We operate with a range where the lower bound
gives at least the desired coverage, but where the higher bound doesn't.
We will take the midpoint of the range, find the cutnodes corresponding to
this value of the threshold, and check if this gives us the desired coverage.
If it does, this becomes the new lower bound, otherwise it becomes the new
upper bound. If the lower and upper bounds are close to each other, we stop
and return the nodes corresponding to the lower bound. This termination
criterion can of course be replaced with something more elaborate.
This can be implemented as follows:
\begin{quote}
{\bf Function} $N(C_0)$
\begin{enumerate}
\item
$S_{low} := 0; S_{high} :=$ largenumber;
$N_c := N(0);$
\item
\label{Label1}
{\bf If} $S_{high} - S_{low} < \delta_S$ \newline
{\bf then} {\bf goto}~\ref{Label5} \newline
{\bf else} $S_{mid} := \frac{S_{low} + S_{high}}{2}$;
\item
$N := N(S_{mid});$
\item
{\bf If} $C(N) < C_0$ \newline
{\bf then} $S_{high} := S_{mid}$ \newline
{\bf else} $S_{low} := S_{mid}$; $N_c := N$;
\item
{\bf Goto}~\ref{Label1};
\item
\label{Label5}
{\bf Return} $N_c$;
\end{enumerate}
\end{quote}
Here $C(N)$ is the coverage on the test set of the specialized grammar
determined by the set of cutnodes $N$.

Actually, we also need to handle the boundary case where no assignment of
cutnodes gives the required coverage. Likewise, the coverages of the upper
and lower bound may be far apart even though the entropy difference is
small, and vice versa. These problems can readily be taken care of by
modifying the termination criterion, but the solutions have been
omitted for the sake of clarity.

In the running example, using the weighted sum of the phrase entropies as
the node entropy, if any threshold value less than 1.08 is chosen,
this will yield any desired coverage, since the single test example of
Figure~\ref{FigTrees} is then covered.

\subsection{Retrieving the specialized rules}

When retrieving the specialized rules, we will match each training example
against the and-or tree. If the current node is a cutnode, we will cut at
this point in the training example. The resulting rules will be the set of
cut-up training examples. A threshold value of say 1.00 in our example will
yield the set of cutnodes $\{n_3,n_4,n_6,n_9\}$ and result in the set of
specialized rules of Figure~\ref{FigSpecRules}.

\begin{figure}
{\small
\begin{verbatim}
1) "S => Det N V Prep NP"

       s_np_vp
         /\
        /  \
np_det_n    vp_vp_pp
   /\          /\
lex  lex   vp_v  pp_prep_np
            |        /\
           lex    lex  NP

2) "S => Pron V NP"

     s_np_vp
       /\
np_pron  vp_v_np
   |        /\
  lex    lex  NP

3) "NP => Det N"

     np_det_n
        /\
     lex  lex

4) "NP => NP Prep NP"

     np_np_pp
        /\
      NP  pp_prep_np
             /\
          lex  NP

5) "NP => Num"

     np_num
        |
       lex
\end{verbatim}
} 
\caption{The specialized rules}
\label{FigSpecRules}
\end{figure}

\begin{figure}
{\small
\begin{verbatim}
6) "S => Det N V NP"

      s_np_vp
        /\
np_det_n  vp_v_np
   /\        /\
lex  lex  lex  NP

7) "S => Pron V Prep NP"

     s_np_vp
       /\
np_pron  vp_vp_pp
   |        /\
  lex   vp_v  pp_prep_np
          |        /\
         lex    lex  NP
\end{verbatim}
} 
\caption{Additional specialized rules}
\label{FigSpecRules1}
\end{figure}

If we simply let the and-or tree determine the set of specialized rules,
instead of using it to cut up the training examples, we will in general
arrive at a larger number of rules, since some combinations of choices in
the and-or tree may not correspond to any training example. If this latter
strategy is used in our example, this will give us the two extra rules of
Figure~\ref{FigSpecRules1}. Note that they not correspond to any training
example.

%
%


\section{ADDITIONAL CONSTRAINTS}

As mentioned at the beginning, the specialized grammar is compiled into
LR parsing tables.
Just finding any set of cutnodes that yields the desired coverage will
not necessarily result in a grammar that is well suited for LR parsing.
In particular, LR parsers, like any other parsers employing a bottom-up
parsing strategy, do not blend well with empty productions.
This is because without top-down filtering, any empty production is
applicable at any point in the input string, and a naive bottom-up
parser will loop indefinitely. The LR parsing tables constitute a type
of top-down filtering, but this may not be sufficient to guarantee
termination, and in any case, a lot of spurious applications of empty
productions will most likely take place, degrading performance.
For these reasons we will not allow learned rules whose RHSs are empty,
but simply refrain from cutting in nodes of the parse trees that do not
dominate at least one lexical lookup.

Even so, the scheme described this far is not totally successful, the
performance is not as good as using hand-coded tree-cutting criteria.
This is conjectured to be an effect of the reduction lengths being
far too short.
The first reason for this is that for any spurious rule reduction to take
place, the corresponding RHS phrases must be on the stack. The likelihood
for this to happen by chance decreases drastically with increased rule
length. A second reason for this is that the number of states visited will
decrease with increasing reduction length. This can most easily be
seen by noting that the number of states visited by a deterministic LR parser
equals the number of shift actions plus the number of reductions, and equals
the number of nodes in the corresponding parse tree, and the longer the
reductions, the more shallow the parse tree.

The hand-coded operationality criteria result in an average rule length of
four, and a distribution of reduction lengths that is such that only 17
percent are of length one and 11 percent are of length two. This is in sharp
contrast to what the above scheme accomplishes; the corresponding figures
are about 20 or 30 percent each for lengths one and two.

An attempted solution to this problem is to impose restrictions on
neighbouring cutnodes. This can be done in several ways; one that has been
tested is to select for each rule the RHS phrase with the least entropy,
and prescribe that if a node corresponding to the LHS of the rule is chosen
as a cutnode, then no node corresponding to this RHS phrase may be chosen
as a cutnode, and vice versa. In case of such a conflict, the node (class)
with the lowest entropy is removed from the set of cutnodes.

We modify the function $N^*$ to handle this:
\begin{quote}
2. {\bf Repeat} $i := i+1$; $N^i := N(N^{i-1}) \setminus B(N^{i-1})$;
\end{quote}
Here $B(N^j)$ is the set of nodes in $N^j$ that should be removed to
avoid violating the constraints on neighbouring cutnodes. It is also
necessary to modify the termination criterion as was done for the
function $N(S_{min})$ above.
Now we can no longer safely assume that the coverage increases with
decreased entropy, and we must also modify the interval-bisection
scheme to handle this. It has proved reasonable to assume that the
coverage is monotone on both sides of some maximum, which simplifies
this task considerably.

\section{EXPERIMENTAL RESULTS}

A module realizing this scheme has been implemented and applied to the
very setup used for the previous experiments with the hand-coded
tree-cutting criteria, see \cite{Samuelsson:94a}.
2100 of the verified parse trees constituted the training set, while 230
of them were used for the test set.
The table below summarizes the results for some grammars of different
coverage extracted using:
\begin{enumerate}
\item
Hand-coded tree-cutting criteria.
\item
Induced tree-cutting criteria where the node entropy was taken to be the
phrase entropy of the RHS phrase of the dominating grammar rule.
\item
Induced tree-cutting criteria where the node entropy was the sum of the
phrase entropy of the RHS phrase of the dominating grammar rule and the
weighted sum of the phrase entropies of the LHSs of the alternative choices
of grammar rules to resolve on.
\end{enumerate}
In the latter two cases experiments were carried out both with and without
the restrictions on neighbouring cutnodes discussed in the previous section.

\vspace{2.5mm}
\noindent
\begin{tabular}{|r|rrrr|rr|}
\hline
\multicolumn{7}{|c|}{Hand-coded tree-cutting criteria}\\
\hline
Coverage	&\multicolumn{4}{|c|}{Reduction lengths (\%)}
&\multicolumn{2}{|c|}{Times (ms)}\\
	&1	&2	&3 	&$\ge 4$			&Ave.	&Med.\\
\hline
90.2 \%	&17.3	&11.3	&21.6	&49.8	&72.6	&48.0\\
\hline
\end{tabular}

\vspace{2.5mm}
\noindent
\begin{tabular}{|r|rrrr|rr|}
\hline
\multicolumn{7}{|c|}{RHS phrase entropy. Neighbour restrictions}\\
\hline
Coverage	&\multicolumn{4}{|c|}{Reduction lengths (\%)}
&\multicolumn{2}{|c|}{Times (ms)}\\
	&1	&2	&3 	&$\ge 4$			&Ave.	&Med.\\
\hline
75.8 \%	&11.8	&26.1	&17.7	&44.4	&128	&38.5\\
80.5 \%	&11.5	&27.4	&20.0	&41.1	&133	&47.2\\
85.3 \%	&14.0	&37.3	&24.3	&24.4	&241	&70.5\\
\hline
\end{tabular}

\vspace{2.5mm}
\noindent
\begin{tabular}{|r|rrrr|rr|}
\hline
\multicolumn{7}{|c|}{RHS phrase entropy. No neighbour restrictions}\\
\hline
Coverage	&\multicolumn{4}{|c|}{Reduction lengths (\%)}
&\multicolumn{2}{|c|}{Times (ms)}\\
	&1	&2	&3 	&$\ge 4$			&Ave.	&Med.\\
\hline
75.8 \%	&8.3	&12.4	&25.6	&53.7	&76.7	&37.0\\
79.7 \%	&9.0	&16.2	&26.9	&47.9	&99.1	&49.4\\
85.3 \%	&8.4	&17.3	&31.1	&43.2	&186	&74.0\\
90.9 \%	&18.2	&27.5	&21.7	&32.6	&469	&126\\
\hline
\end{tabular}

\vspace{2.5mm}
\noindent
\begin{tabular}{|r|rrrr|rr|}
\hline
\multicolumn{7}{|c|}{Mixed phrase entropies. Neighbour restrictions}\\
\hline
Coverage	&\multicolumn{4}{|c|}{Reduction lengths (\%)}
&\multicolumn{2}{|c|}{Times (ms)}\\
	&1	&2	&3 	&$\ge 4$			&Ave.	&Med.\\
\hline
75.3 \%	&6.1	&11.7	&30.8	&51.4	&115.4	&37.5\\
\hline
\end{tabular}

\vspace{2.5mm}
\noindent
\begin{tabular}{|r|rrrr|rr|}
\hline
\multicolumn{7}{|c|}{Mixed phrase entropies. No neighbour restrictions}\\
\hline
Coverage	&\multicolumn{4}{|c|}{Reduction lengths (\%)}
&\multicolumn{2}{|c|}{Times (ms)}\\
	&1	&2	&3 	&$\ge 4$			&Ave.	&Med.\\
\hline
75 \%	&16.1	&13.8	&19.8	&50.3	&700	&92.0\\
80 \%	&18.3	&16.3	&20.1	&45.3 	&842	&108\\
\hline
\end{tabular}
\vspace{2.5mm}

With the mixed entropy scheme it seems important to include the restrictions
on neighbouring cutnodes, while this does not seem to be the case with the
RHS phrase entropy scheme.
A potential explanation for the significantly higher average parsing times
for all grammars extracted using the induced tree-cutting criteria is that
these are in general recursive, while the hand-coded criteria do not allow
recursion, and thus only produce grammars that generate finite languages.

Although the hand-coded tree-cutting criteria are substantially better than
the induced ones, we must remember that the former produce a grammar that in
median allows 60 times faster processing than the original grammar and parser
do. This means that even if the induced criteria produce grammars that are a
factor two or three slower than this, they are still approximately one and a
half order of magnitude faster than the original setup. Also, this is by no
means a closed research issue, but merely a first attempt to realize the
scheme, and there is no doubt in my mind that it can be improved on most
substantially.

\section{SUMMARY}

This article proposes a method for automatically finding the appropriate
tree-cutting criteria in the EBG scheme, rather than having to hand-code
them. The EBG scheme has previously proved most successful for tuning a
natural-language grammar to a specific application domain and thereby
achieve very much faster parsing, at the cost of a small reduction in
coverage.

Instruments have been developed and tested for controlling the coverage
and for avoiding a large number of short reductions, which is argued to
be the main source to poor parser performance. Although these instruments
are currently slightly too blunt to enable producing grammars with the
same high performance as the hand-coded tree-cutting criteria, they can
most probably be sharpened by future research, and in particular refined
to achieve the delicate balance between high coverage and a distribution
of reduction lengths that is sufficiently biased towards long reductions.
Also, banning recursion by category specialization, i.e.~by for example
distinguishing {\it NP}s that dominate other {\it NP}s from those that do
not, will be investigated, since this is believed to be an important
ingredient in the version of the scheme employing hand-coded tree-cutting
criteria.

\section*{ACKNOWLEDGEMENTS}

This research was made possible by the basic research programme at the
Swedish Institute of Computer Science (SICS). I wish to thank Manny Rayner
of SRI International, Cambridge, for help and support in matters pertaining
to the treebank, and for enlightening discussions of the scheme as a whole.
I also wish to thank the NLP group at SICS for contributing to a very
conductive atmosphere to work in, and in particular Ivan Bretan for valuable
comments on draft versions of this article. Finally, I wish to thank the
anonymous reviewers for their comments.

\end{document}